\documentclass{article}

\usepackage{booktabs}
\usepackage{siunitx}
\usepackage{bm,bbm}
\usepackage{amssymb,amsmath}
\usepackage{hyperref}
\usepackage{graphicx}

\usepackage{fourier}

\usepackage{color}

\title{Classification of Complex Wishart Matrices with a Diffusion-Reaction System guided by Stochastic Distances}

\author{Luis Gomez\footnote{CTIM.\ Dpto. de Ingenier\'{\i}a Electr\'onica y Autom\'atica,
Universidad de Las Palmas de G.C.,
Campus de Tafira, 35017, Spain,
luis.gomez@ulpgc.es}
\and Luis Alvarez\footnote{CTIM.\ Dpto.\ de Inform\'atica y Sistemas,
Universidad de Las Palmas de G.C.,
Campus de Tafira, 35017, Spain,
lalvarez@ctim.es}
\and Luis Mazorra\footnote{lmazorra@ctim.es}
\and Alejandro C.\ Frery\footnote{Laborat\'orio de Computa\c c\~ao Cient\'ifica e An\'alise Num\'erica,
Universidade Federal de Alagoas,
Av.\ Lourival Melo Mota, s/n,
57072-900 Macei\'o -- AL, Brazil,
acfrery@gmail.com}
}

\begin{document}

\begin{abstract}
We propose a new method for PolSAR (Polarimetric Synthetic Aperture Radar)  imagery classification based on stochastic distances in the space of random matrices obeying complex Wishart distributions. 
Given a collection of prototypes $\{Z_m\}_{m=1}^M$ and a stochastic distance $d(.,.)$, we classify any random matrix $X$ using two criteria in an iterative setup.
Firstly, we associate $X$ to the class which minimizes the weighted  stochastic distance $w_md(X,Z_m)$, where the positive weights $w_m$ are computed to maximize the class discrimination power. 
Secondly, we improve the result by embedding the classification problem into a diffusion-reaction partial differential system where the diffusion term smooths the patches within the image, and the reaction term tends to move the pixel values towards the closest class prototype. 
In particular, the method inherits the benefits of speckle reduction by diffusion-like methods. 
Results on synthetic and real PolSAR data show the performance of the method. \\
\textbf{Keywords:}
PolSAR images, Classification , Diffusion-Reaction system
\end{abstract}

\maketitle

\section{Introduction}

Classification is one of the most important techniques for image analysis.
It aims at mapping each pixel into a class, so it transforms values into information.

Classification can be performed using a variety of sources, mostly the spectral information (the observed value in each pixel), spatial or contextual information, and ancillary data (ground truth, for instance).
The latter is usually only available in very restricted areas, from which training samples can be obtained.

The simplest available classification techniques rely only on pixel-wise information, i.e., on the observation in each coordinate: Isodata, Parallelepiped and Pointwise Maximum Likelihood are examples of these methods.
Arguably, the most successful techniques exploit both the spectral information and the context.
This is mostly due to the fact that images exhibit a great deal of spatial redundancy, i.e., spatially neighboring pixels tend to be alike.

As an example of contextual classification one should mention techniques based on Markovian models. 
Geman and Geman~\cite{geman84} posed the classification process as an estimation problem and, as such, proposed a number of estimators and algorithms.
These techniques rely on variations of the following idea:
the class $m$ in each coordinate should satisfy a criterion that, simultaneously, optimizes the pointwise spectral evidence and a contextual measure of smoothness, for example maximizing
\begin{equation}
\lambda f_m (z(i,j)) + (1-\lambda) N(m, \partial_{ij}),
\label{eq:MAPMarkovian}
\end{equation}
where $\lambda\in[0,1]$ is the relative weight of the spectral evidence over the context,
$ f_m (z(i,j)) $ is the likelihood of the observation $z(i,j)$ in coordinate $(i,j)$ with respect to the model characterized by the probability density function $ f_m$, 
$m\in\{1,\dots,M\}$ is one of the $M$ possible classes, and
$N(m, \partial_{ij})$ is a nondecreasing function on the number of neighbors of coordinate $(i,j)$, denoted by $\partial_{ij}$, that have been classified as $m$.
If $\lambda=0$, only the context is relevant and the most frequent class in the neighbours is the optimal solution.
If $\lambda=1$, only the radiometric pointwise information is relevant, and the maximum likelihood estimator (assuming independent observations) is the optimal solution.
Specialized algorithms, such as the ICM (Iterated Conditional Modes), MPM (Maximum Posterior Modes), and Simulated Annealing are required to obtain (often approximate) solutions for intermediate values of $\lambda$; cf.\ Ref.~\cite{FreryCorreiaFreitas:ClassifMultifrequency:IEEE:2007}.
Eq.~\eqref{eq:MAPMarkovian} makes it clear that the context, in this approach, is incorporated by means of the neighbouring classes.

The aim of this work is proposing a new classification procedure based on both context and radiometric information,
but without the use of classes as descriptors of the former.
For the latter, we tackle the problem of classifying Polarimetric Synthetic Aperture Radar (PolSAR) imagery.

According to Ref.~\cite{LeePottier2009PolarimetricRadarImaging}, the early developments of imaging radar aimed at the characterization of aircraft targets.
Spatial resolution was a fundamental issue with these images;
it depends directly on the dimensions of the antenna, more precisely, on its \textit{aperture}.
Deployable antennas would yield unacceptable spatial resolutions of the order of hundreds of metres, even kilometres.
Higher resolutions can be obtained by controlling the illumination, forming longer \textit{synthetic} (as opposed to physical) antennas, yielding the broad area of SAR -- \textit{Synthetic Aperture Radar}.
 
The main differences between SAR sensors and sensors that operate in the visible spectrum are
\begin{description}
\item[Wavelength:] The former operate in the microwave section of the electromagnetic spectrum.
\item[Activeness:] SAR sensors carry their own source of illumination.
\item[Coherence:] SAR sensors are able to record the relative phase of the emitted and incidence signals.
\end{description}

These characteristics lead to a number of interesting and, oftentimes, challenging properties, among them:
\begin{itemize}
\item SAR sensors are little affected by adverse weather conditions, such as fog, rain, smog, etc., and they provide images regardless the presence of daylight.
\item The return is mostly sensitive to the target dielectric properties, and to its geometry at the wavelength scale, which typically ranges from millimeters to meters.
\item SAR images, after being processed, can be adequately described by a multiplicative rather than additive model for the data.
\end{itemize}

Polarimetric SAR (PolSAR) extends the imaging ability of single SAR imaging.
A PolSAR sensor transmits a linearly polarized signal and records two orthogonal polarizations of the returned signal.
Such multidimensional information allows deeper investigation of the scattering mechanisms of the targets under study.
Fully PolSAR images record the four possible combinations of the signal according to the polarization of the transmitting and receiving antennas.
The first practical fully PolSAR sensor, the L-band AIRSAR, was developed by the Jet Propulsion Laboratory (JPL) in 1985.

On the one hand, the deterministic approach has been used with success in the classification of PolSAR imagery, usually at the expense of using additional information.
For instance, Shi et al.~\cite{GraphEmbeddingPolSARClassification} perform Manifold Learning on layers of features derived from polarimetric decompositions, while
Negri et al.~\cite{SVNContextualClassification} use neighbouring information and Support Vector Machines.

On the other hand, statistical approaches able to cope with the departure from the usual Gaussian additive data formation provide an attractive framework for PolSAR image classification.
Among others, Refs.~\cite{FreitasFreryCorreia:Environmetrics:03,lee94,SimulationCorrelatedPolSAR} provide a comprehensive account of the models, and of the relations among them, that have been used to describe polarized and fully polarimetric SAR images.

The classification of PolSAR imagery is an important research topic.
Among the ones that use stochastic models for the fully polarimetric information, one should mention the pioneering work of Lee et al.~\cite{leegrunes94}, who present the multivariate complex Wishart distribution 
and some of its properties for the Remote Sensing community.
Other techniques have been proposed as, for instance, the minimization of stochastic distances between segments of data and prototypes~\cite{ClassificationPolSARSegmentsMinimizationWishartDistances}, and the use of spectral and contextual information by means of the ICM algorithm under the Potts model as prior distribution~\cite{FreryCorreiaFreitas:ClassifMultifrequency:IEEE:2007}.
These works use the data as available to the users, differently from another approach, namely the classification based on the $H/\alpha$ decomposition~\cite{PolarimetricDecomposition1999}.
It is noteworthy that the latter work presents a deterministic and unsupervised technique, while the others rely on the availability of prototypes and stochastic models.

Diffusion-reaction equations have been successfully applied in many different contexts, e.g. Chemistry and Ecology. 
In the field of image processing, diffusion-reaction equations have been applied for instance in quantization~\cite{EsAl97}, filtering and segmentation~\cite{weickert1998anisotropic}, and in medical applications~\cite{Ko2007}.

The general shape of a diffusion-reaction partial differential equation is given by 
\begin{equation}
\frac{\partial u}{\partial t}= \mathcal{L}(u) +f(u),
\label{DRE}
\end{equation}
where $u(t,x,y)$ represents the solution of the equation at instant $t$ and coordinates $(x,y)$, $\mathcal{L}(u)$ is the diffusion term given, typically, by an elliptic second order differential operator, for instance the Laplacian operator. 
The reaction term is $f(u)$, whose shape strongly depends on the intended application. 
Roughly speaking, the reaction term tends to move the solution $u(t,x,y)$ of the equation towards the nearest stable equilibrium state associated to the ordinary differential equation (ODE) $u'=f(u)$ ($u_m$ is an stable equilibrium state of $u'=f(u)$ if $f(u_m)=0$ and $f'(u_m)<0$). 
Usually, $u(0,x,y)$ represents the observed data and $u(t,x,y)$ represents its evolution under the action of the differential equation where the diffusion and reaction terms are in competition. 
On the one hand, the diffusion term tends to smooth the solution taking into account the neighbors values and, on the other hand, the reaction term tends to move the solution towards some asymptotic values.

In this work we address a difficult problem:
the classification of PolSAR images.
Our approach is based on weighted stochastic distances and a diffusion-reaction partial differential equations system which imposes a certain level of spatial smoothness. 
This work is an extension of Ref.~\cite{GAMF_IWOBI2015}, where we proposed a diffusion-reaction system, but using the Euclidean distance to perform PolSAR image classification. 
We will show that the use of weighted stochastic distances, instead of the Euclidean distance, improves significantly the classification results.


The rest of the article is organized as follows: 
Section~\ref{correction:model} deals with the statistical description of PolSAR data, including expressions of stochastic distances between models.
Section~\ref{Sec:Methodology} is devoted to the classification of these data using weighted stochastic distances between complex Wishart distributions. 
Section~\ref{Sec:DiffussionReaction} deals with the discretization of the diffusion-reaction differential system. 
Section~\ref{Sec:Experimental} presents results on the classification of simulated and real PolSAR data, and, finally, 
in Section~\ref{Sec:Conclusion} some conclusions are drawn.

\section{The Scaled Complex Wishart distribution}\label{correction:model}

This section is based on the work by Nascimento et al.~\cite{BiasCorrectionModifiedProfileLikelihoodWishartDistribution}.

\subsection{Characterization}

The polarimetric coherent information associates, in each frequency of operation, to each pixel  a 2$\times$2 complex matrix with entries
$S_\text{VV}$, $S_\text{VH}$, $S_\text{HV}$, and $S_\text{HH}$, 
where $S_{ij}$ is the backscattered signal 
for the $i$th transmission and $j$th reception linear polarization, 
$i,j=\text{H},\text{V}$.
Under mild conditions, $S_\text{HV}=S_\text{VH}$,
and the scattering matrix can be simplified into the three-component complex vector
$\bm{s}=\big[
S_\text{VV}
\;\;
\sqrt{2}\,S_\text{VH}
\;\;
S_\text{HH}
\big]^\top
$, where ${}^\top$ denotes vector transposition.
This random vector 
can be
modeled by 
the zero-mean multivariate complex Gaussian distribution~\cite{Goodman63b}.

Different targets are characterized by different variances and, thus, are hard to discern visually.
In order make such difference noticeable, and to improve the signal-to-noise ratio,
$L$ ideally independent measurements of the same target are obtained while processing the raw data.
This quantity is the number of looks.
These observations are used to produce the ``multilook sample covariance matrix format''
$$
\bm{Z}
=
\frac{1}{L}
\sum_{i=1}^L 
\bm{s}_i \bm{s}_i^{*},
$$
where the superscript ${*}$ 
represents the complex conjugate transpose of a vector, and
$\bm{s}_i$, \mbox{$i=1,2,\ldots,L$}, are the $L$ scattering vectors.
Assuming that these vectors are independent, $\bm{Z}$ is an
Hermitian positive definite matrix,
and it
follows a scaled complex Wishart distribution~\cite{EstimationEquivalentNumberLooksSAR}. 
Having $\bm{\Sigma}$ and $L$ as parameters, 
the scaled complex Wishart distribution is characterized by the following probability density function
\begin{equation}
\label{eq:denswishart}
f_{\bm{Z}}(\bm{z};\bm{\Sigma},L) 
=
\frac{L^{3L}|\bm{z}|^{L-3}}{|\bm{\Sigma}|^L \Gamma_3(L)} 
\exp\big(
-
L
\operatorname{tr}(\bm{\Sigma}^{-1}\bm{z})
\big)
,
\end{equation}
where 
$\Gamma_3(L)=\pi^{3}\prod_{i=0}^{2}\Gamma(L-i)$ 
for $L\geq 3$, 
$\Gamma(\cdot)$ is the gamma function, 
$\operatorname{tr}(\cdot)$ represents the trace operator, 
$|\cdot|$ denotes the determinant operator, 
$\bm{\Sigma}$ is the covariance matrix
associated to~$\bm{s}$, \mbox{$\Sigma=\operatorname{E}\bigl(\bm{s}\bm{s}^{*}\bigr)$},
where $\operatorname{E}(\cdot)$ is the 
expectation.
The first moment
of $\bm{Z}$
satisfies
$\operatorname{E}(\bm{Z})=\bm{\Sigma}$.
We denote
$\bm{Z}\sim \mathcal W(\bm{\Sigma},L)$ 
to indicate that $\bm{Z}$
follows the scaled complex Wishart distribution.
As can be seen in~\cite{AnalyticExpressionsStochasticDistancesBetweenRelaxedComplexWishartDistributions}, this distribution is able to accommodate an arbitrary number of polarimetric components.

\subsection{Parameter estimation}\label{ref12}

Let $\{\bm{Z}_1,\bm{Z}_2,\ldots,\bm{Z}_N\}$ 
be a random sample from 
$\bm{Z}\sim \mathcal W(\bm{\Sigma},L)$. 
The ML estimators for $\bm{\Sigma}$ and $L$, 
namely $\widehat{\bm{\Sigma}}_{\text{ML}}$ and $\widehat{L}_{\text{ML}}$, 
respectively, 
are quantities that maximize 
the log-likelihood function associated to 
the Wishart distribution,
which is given by
\begin{equation}
\begin{split}
{\ell}(\bm{\theta})
=
&
3NL\log L
+
(L-3)\sum_{k=1}^N
\log|\bm{Z}_k|
-
LN\log|\bm{\Sigma}| 
\\
&
-3N
\log\pi
-N
\sum_{i=0}^{2}
\log\Gamma(L-i)
-NL
\operatorname{tr}(\bm{\Sigma}^{-1}\bar{\bm{Z}})
,
\end{split}
\label{likelihood}
\end{equation}
where
$\bar{\bm{Z}}=N^{-1}\,\sum_{k=1}^N\,\bm{Z}_k$,
$\bm{\theta}=\big[\bm{\sigma}^\top \;\; L \big]^\top$,
$\bm{\sigma}=\operatorname{vec}(\bm{\Sigma})$,
and $\operatorname{vec}(\cdot)$ 
is the column stacking vectorization operator.
Frery et al.~\cite{EntropyBasedStatisticalAnalysisPolSAR} showed that
$\widehat{\bm{\Sigma}}_\text{ML}$
is given by the sample mean
\begin{equation}
\widehat{\bm{\Sigma}}_\text{ML}=\bar{\bm{Z}},
\label{Sigma_estimator}
\end{equation}
and that $\widehat{L}_{\text{ML}}$ 
satisfies
the following non-linear equation
\begin{align}
\label{eqscore1} 
3\log\widehat{L}_\text{ML}
+
\frac{1}{N}
\sum_{k=1}^N 
\log|\bm{Z}_k|
-
\log|\bar{\bm{Z}}|
-
\psi_3^{(0)}(\widehat{L}_{\text{ML}})
=
0,
\end{align}
where $\psi_3^{(0)}(\cdot)$ is the 
zeroth order term of the $v$th-order
multivariate polygamma function given by
$$
\psi_3^{(v)}(L)=\sum_{i=0}^{2} \varphi^{(v)}(L-i),
$$
where $\varphi^{(v)}(\cdot)$ is the ordinary polygamma function expressed by
$$
\varphi^{(v)}(L)=\frac{\partial^{v+1} \log\Gamma(L)}{\partial L^{v+1}},
$$
for $v\geq 0$, and $\varphi^{(0)}(\cdot)$ is the digamma function.

Nascimento et al.~\cite{BiasCorrectionModifiedProfileLikelihoodWishartDistribution} noted that $\widehat{L}_{\text{ML}}$ is biased, and proposed a number of correction techniques obtained subtracting an estimator of the bias from the ML estimator.
Using their results, we employ the Box-Snell estimator of the bias, given by
\begin{equation}
\widehat{B}(L) = \frac{9}{2NL\Big[ \psi^{(1)}_3(L)-\frac{3}{L} \Big]} - 
\frac{\frac{3}{2L}+\psi^{(2)}_3(L)}{2N\Big[ \psi^{(1)}_3(L) - \frac{3}{L} \Big]}.
\label{eq:BiasLML}
\end{equation}
We first compute $\widehat{L}_{\text{ML}}$ by solving~\eqref{eqscore1}, 
then we evaluate $\widehat{B}(\widehat{L}_{\text{ML}})$ with~\eqref{eq:BiasLML} and, finally, we use the
first-order bias-corrected estimator $\widetilde{L} = \widehat{L}_{\text{ML}} - \widehat{B}(\widehat{L}_{\text{ML}})$.

\subsection{Stochastic distances between Wishart laws}

Frery et al.~\cite{AnalyticExpressionsStochasticDistancesBetweenRelaxedComplexWishartDistributions},
while seeking for contrast measures between samples, 
derived expressions for the stochastic distances between pairs of Wishart laws with (possibly) different covariance matrices $\boldsymbol{\Sigma}_1$ and $\boldsymbol{\Sigma}_2$, and same number of looks $L$.
They obtained the Kullback-Leibler, Hellinger, Bhattacharyya and R\'enyi (of order $\beta$) distances.

The two first are of interest in this work, and they are given, respectively, by
\begin{align}
d_\text{KL}(\boldsymbol{\Sigma}_1, \boldsymbol{\Sigma}_2) &= L\bigg[\frac{\operatorname{tr}(\boldsymbol{\Sigma}_1^{-1}\boldsymbol{\Sigma}_2+\boldsymbol{\Sigma}_2^{-1}\boldsymbol{\Sigma}_1)}{2}-3\bigg],
\label{KL-distance}\\
\text{and} \nonumber\\
d_\text{H}(\boldsymbol{\Sigma}_1, \boldsymbol{\Sigma}_2) &= 1 - \Bigg[
\frac{
\big( 
\boldsymbol{\Sigma}_1^{-1} + \boldsymbol{\Sigma}_2^{-1}
\big)^{-1}
}{
2\sqrt{|\boldsymbol{\Sigma}_1| |\boldsymbol{\Sigma}_2|}
}
\Bigg]^L . \label{H-distance}
\end{align}
Notice that both distances depend on two very simple operations: the determinant and the inverse.

The two other distances, namely R\'enyi and Bhattacharyya, were not used because the former depends on an extra parameter ($\beta$), and the latter is a simple increasing transformation of the Hellinger distance, namely $d_{\text{B}}=-\log(1-d_{\text{H}})$, so it adds little to this study.

\section{Complex Wishart matrices classification using weighted stocastic distances}\label{Sec:Methodology}

In this paper we address a multiclass classification problem in $\mathcal{A}$, the space (or cone) of the
of Hermitian positive definite matrices.
All elements in $\mathcal A$ are assumed to follow a scaled complex Wishart distribution, as defined by the density given in Eq.~\eqref{eq:denswishart}.

We assume there are $M$ classes and, for each class $m\in\{1,\ldots, M\}$, 
we denote by $Z_m$ its prototype, given by a random matrix in  $\mathcal{A}$ that follows the scaled complex Wishart distribution $\mathcal{W}(\Sigma_{m},L_m)$. 
For each point of a PolSAR image we have a random matrix $X$ which is assumed to belong to one of the predefined classes. 

A basic classification procedure is to assign each $X\in\mathcal A$ to the class $m$ for which
$X$ is closest to $Z_m$, in a stochastic distance sense.
This pointwise procedure is prone to errors due to the variability of the observations.
In order to improve this basic classification procedure, we introduce a collection of weights $\{w_m\}$ such that
$X$ is associated to the class $m_{\min}(X)\in\{1,\ldots,M\}$ satisfying 
\begin{equation}
m_{\min}(X)=\arg\min_{m}w_m  d(X,Z_{m}),
\label{m_min}
\end{equation}
i.e., 
among the $M$ possible classes, $m_{\min}(X)$ is the 
class whose prototype $Z_m$ is 
closest to $X$ with respect to a distance $d$ weighted by a factor that depends on the class.
If $w_m=\infty$ for any class, that class is forbidden; if $w_m=0$, the classification into $m$ is mandatory.
In the following subsection we will propose a procedure to estimate $\{w_m\}$  in order to maximize the discrimination power of the classes. 

These multiplicative factors have the effect of modifying the number of looks $L_m$ of each class
either directly, as in the Kullback-Leibler distance, Eq.~\eqref{KL-distance},
or as a first order approximation in the Hellinger  distance, Eq.~\eqref{H-distance}.
Their purpose is to exert control on the relative importance each class has on the procedure.



Each of the $M$ prototypes is built by selecting a ROI (region of interest) that will serve as ground truth or reference. 
Each ROI is split in two disjoint sets of equal size by simple random sampling without replacement. 
The first set is used as training data, while the second is employed as test data. 
For each class, the training samples are used  to estimate $(\Sigma_m,L_m)$ by improved maximum likelihood, respectively, as described in Section~\ref{correction:model}\ref{ref12}.
Denote by $M_m$ the number of pixels in the training set of class $m$, $1\leq m\leq M$, and its pixels values by $Z_m^k$, $1\leq k\leq M_m$.

The weights $w_m$ are computed such that they maximize the discrimination power of the stochastic distance.
This is done by using an energy optimization strategy, with the aid of the following energy function
%
\begin{equation}
\sum_{m=1}^{M}\frac{1}{M_m}\sum_{k=1}^{M_{m}}\sum_{m^{\prime}\neq
m}\phi\big(  w_{m}d(Z_{m}^{k},Z_{m})-w_{m^{\prime}}d(Z_{m}^{k%
},Z_{m^{\prime}})\big),
\label{energy}
\end{equation}
where 
$$
\phi(s)=\frac{s}{1+\lambda|s|},
$$
and $\lambda>0$ is a parameter.  
In order to simplify the optimization problem we assume that \mbox{$\sum_m w_m=1$}. 
Roughly speaking, by minimizing Eq.~\eqref{energy} we make the observations $Z^{k}_m$ be as close as possible to $Z_m$, and as far as possible to any other class representative $Z_{m'}$.  
The minimum of the above energy function is found using a gradient descent type algorithm. 
In all the experiments showed in this paper, $w_m$ is initialized as $1/M$ and $\lambda$ is fixed to $1$. 


\section{Diffusion-reaction differential system}\label{Sec:DiffussionReaction}

This section follows the ideas introduced in Ref.~\cite{GAMF_IWOBI2015}, but using weighted stochastic distances instead of the Euclidean distance. 
In order to introduce a smoothing procedure in the classification step, we embed the classification problem in a  Diffusion-Reaction differential system. 

We denote by $\Sigma(0,x,y)$ the initial covariance matrix provided by the PolSAR image, 
and by $\Sigma(t,x,y)$ its evolution under the action of the diffusion-reaction system. 
Let $Z(t,x,y)$ be the family of Wishart random matrices associated to $\Sigma(t,x,y)$. 
The diffusion-reaction system we propose is given by
\begin{equation}
\frac{\partial\Sigma}{\partial t}=\alpha\Delta\Sigma + 
\big(
\min_{m}w_m d(Z,Z_{m}) - 
\min_{m\neq m_{\min}(Z)}w_m d(Z,Z_{m})\big) \big(\Sigma-\Sigma_{m_{\min}(Z)}
\big),\label{PDE}%
\end{equation}
where $\Delta\Sigma$ represents the spatial Laplacian differential operator applied on each component of the matrix $\Sigma(t,x,y)$, 
$\{Z_{m}\}$ represents distributions that describe the classes present in the PolSAR image, and 
$\alpha\geq 0$ is a weight parameter which balances the influence of the diffusion and reaction terms,
and $\Sigma_{m_{\min}}(Z)$ is obtained using eq.~\eqref{m_min}.
The diffusion term, given by the Laplacian operator tends to smooth the covariance matrix $\Sigma(t,x,y)$, and the reaction term, given by the remainder of the equation, tends to move
$\Sigma(t,x,y)$ towards the nearest covariance matrix $\Sigma_{m}$ according to the weighted stochastic distance $w_md(Z(t,x,y),Z_{m})$. 

\subsection{Diffusion-reaction system discretization}

To discretize equation~\eqref{PDE} we denote by $Z_{i,j}^{n}$ the
approximation $Z(n \delta t, i h, j h)$, and by $\Sigma_{i,j}^{n}$
its associated covariance matrix, where $i,j\in \mathbbm N$, and $\delta t$ and $h$ are the
discretization values. 
We propose the following two-step algorithm where, 
in the first step, we discretize the contribution of the diffusion part of the equation and, then, in the second step, we discretize the contribution of the reaction part of the equation taking into account that, locally, using a linear approximation of the reaction term, the solution of the equation is given by an exponential function:
\begin{align*}
\label{RD_scheme}
\text{Step 1:}\\
\Sigma_{i,j}^{n\prime}&=\Sigma_{i,j}^{n}%
+\alpha\delta t\frac{\Sigma_{i+1,j}^{n}+\Sigma_{i-1,j}^{n}+\Sigma_{i,j+1}%
^{n}+\Sigma_{i,j-1}^{n}-4\Sigma_{i,j}^{n}}{h^{2}} 	\\
\text{Step 2:}\\
\Sigma_{i,j}^{n+1}&=\Sigma_{m_{\min}(Z_{ij}%
^{n\prime})}+e^{\delta t\Big(  
\underset{m}{\min}\ w_m d(Z_{i,j}^{n\prime}%
,Z_{m})-\underset{m\neq m_{\min}(Z_{ij}^{n\prime})}{\min} w_m d(Z_{ij}^{n\prime
},Z_{m})
\Big)  }(\Sigma_{ij}^{n\prime}-\Sigma_{m_{\min}(Z_{ij}^{n\prime})})
\end{align*}
where $\Sigma_{i,j}^{n\prime}$ represents the covariance evolution using only
the diffusion operator,
 and $Z_{ij}^{n\prime}$ its associated random matrix. 
 Notice that the first step incorporates the contextual evidence.

We observe that if $1-4\alpha\delta t/h^{2}\geq0$, then $\Sigma_{i,j}^{n\prime}$
is a convex combination of $\Sigma_{i,j}^{n},$ $\Sigma_{i+1,j}^{n}$,
$\Sigma_{i-1,j}^{n}$, $\Sigma_{i,j+1}^{n}$, $\Sigma_{i,j-1}^{n}\in\mathcal{A}$,
and, in particular, $\Sigma_{i,j}^{n\prime}\in\mathcal{A}$. 
Moreover since
$\Sigma_{i,j}^{n+1}$ is a convex combination of $\Sigma_{m_{\min}(Z_{ij}^{n}%
)}$ and $\Sigma_{i,j}^{n\prime}$, then $\Sigma_{i,j}^{n+1}\in\mathcal{A}$.

That is, if the original covariance matrices $\big\{
\Sigma_{i,j}^{0}\big\}  _{i,j\in N}$  provided by the original PolSAR image
belong to $\mathcal{A}$, then $\left\{  \Sigma_{i,j}^{n}\right\}  _{i,j\in
N}\subset\mathcal{A}$ for all $n>0$ and, therefore, the associated random
matrices $Z_{i,j}^{n}$ are well defined. 
Notice that the above discretization scheme depends on three parameters: $\alpha$, $\delta t$ and  $h$. 
However, looking at the shape of the scheme, we point out that it depends just on the values $\alpha\delta t/h^2$ (Step~1), and on $\delta t$ (Step~2); therefore, 
without loss of generality, we fix $h=1$.  
The only parameters that have influence on the results are $\alpha$ and $\delta t$. 
As explained above, in order to have a well-defined procedure with solutions in $\mathcal{A}$, 
these parameters must satisfy the condition $1-4\alpha\delta t\geq 0$; 
$\delta t$ represents the time discretization step in the evolution equation. 

The smaller $\delta t$ is, the better becomes the approximation of the above scheme to the actual continuous  solution of the differential system given by eq.~\eqref{PDE}. 
In practice, as far as $\delta t$ is small enough we do not expect a strong influence of $\delta t$ in the results. 
In practice, the most important parameter is $\alpha$. 
The larger $\alpha$ is, the stronger the smoothing effect of the differential system given in eq.~\eqref{PDE} becomes. 

In all the experiments presented in this paper, the discretization parameters have been fixed to $\delta t=0.01$ and $\alpha=0.5$. 
Different choices and combinations are possible, and a detailed comparison study will be performed.

\section{Experimental Setup}\label{Sec:Experimental}

We compare the performance of the proposed methodology by classifying both simulated and real PolSAR images with respect to the following techniques
\begin{enumerate}
\item Maximum likelihood (ML) under the Wishart model: each observation is assigned to the class whose density, as expressed in Eq.~\eqref{eq:denswishart}, is maximized.
\item Euclidean distance (ED): each observation is classified into the class which minimizes the Euclidean distance between the observation and its prototype.
\item Hellinger distance (HD): each observation receives the label of that class which minimizes this distance to its prototype; cf. Eq.~\eqref{H-distance}.
\item\label{item:KL} Kullback-Leibler distance (KL):  same as above, but using Eq.~\eqref{KL-distance}.
\item KL distance and optimized weights  (KL+OW): same as in~\ref{item:KL}, but using the optimized weights computed minimizing the energy functional~\eqref{energy}.
\item Diffusion reaction equation with KL+OW  (DR+KL+OW+$n$): The reaction diffusion equation is solved using $n$ iterations of the iterative discretization scheme, then each resulting observation is classified using KL+OW. 
\end{enumerate}

The visualization of PolSAR images requires stipulating a projection of the elements of $\mathcal A$ into the unitary cube, then mapping these three components to the red, green and blue components of a colour image (the RGB representation).
Lee and Pottier~\cite{LeePottier2009PolarimetricRadarImaging} discuss a number of such projections.
We adopt here a simple visualization technique that emphasizes the quality of the classifications obtained.
We assign a unique colour to each class representative $\Sigma_m$, then, we assign a colour to any $\Sigma_{i,j} \in \mathcal{A}$ using a linear interpolation procedure based on the Euclidean distance bewteen $\Sigma_{i,j}$ and $\Sigma_m$. 
The smaller the Euclidean distance between $\Sigma_{i,j}$ and $\Sigma_m$, the larger the weight of the colour component of class $m$ is in the interpolation procedure. 

\subsection{Simulated Image}

In the first experiment we use a $300\times300$ pixels phantom image with three classes generated using three Wishart distributions.
The parameters used for the simulation are those estimated in the training samples used in the experiment discussed in the next section.
These training samples come from the ocean, forest and urban classes identified in Fig~\ref{fig:SF_01}.
Each deviate from a Wishart law is obtained by simulating entries of the complex matrix entries, which obey complex Gaussian observations, then forming the sample covariance matrix and adding as many of them as the (integer) number of looks.
This requires stipulating the number of looks, that in our case was $L=4$ for all classes, and the variance and covariances of the these entries~\cite{Goodman63b}.

Table~\ref{table:phantom} presents the overall accuracy of classifications produced by the techniques here considered.
The optimized weights, obtained through the minimization of the energy function presented in eq.~\eqref{energy}, were $\omega=(0.23,0.50,0.28)$. 
In all cases, the equivalent number of looks was informed $L=4$ and fixed for the three classes. 
For each class, the technique which provides the worst classification is  used as baseline for computing the improvement each technique provides. 
These improvements are informed as percentages between parentheses.

\begin{table}[hbt]
\centering
\begin{tabular}[c]{cr*4{l}}\toprule
\textbf{Method} & \multicolumn{4}{c}{\textbf{Accuracy (Improvement) by Class}} & 
\hspace{-3mm} \begin{tabular}{c} \textbf{CPU}\\ \textbf{time} \end{tabular}\\  
\cmidrule(lr{.75em}){2-5}
& Class 1 (Ocean) & Class 2 (Forest)& Class 3 (Urban)&  & \hspace{0mm} (s)\\ \midrule
ML & $100$ \hspace{10mm} & $98.5$ ($77.6\%$) & $96.6$ ($87.3\%$) & & 0.170 \\
ED & $100$ \hspace{10mm}  & $93.3$ (baseline) & $83.1$ ($36.7\%$) &  & 0.005\\
HD & $100$ \hspace{10mm}  & $99.9$ ($98.5\%$) & $82.8$ ($35.6\%$) & & 0.200 \\
KL & $100$ \hspace{10mm}  & $99.7$ ($95.5\%$) & $73.3$ (baseline) &  &0.020 \\
KL+OW & $100$ \hspace{10mm}  & $96.6$ ($49.3\%$) & $93.1$ ($74.2\%$) & &1.210\\
DR+KL+OW+$50$ & $100$ \hspace{10mm}  & $99.7$ ($95.5\%$) &  $100$ ($100\%$) & & 3.630\\
\bottomrule
\end{tabular}
\caption{\label{table:phantom} Phantom classification scores using the test data.}
\end{table}

All techniques classified all points of Class~1 properly.
The worst classifications for classes~2 and~3 were produced, respectively, by the Euclidean and by the Kullback-Leibler distances.
Class 3 presents the largest variability, hence improvement, in the results. 
The Hellinger distance is the second best technique with a marginal improvement of $35.6\%$ over the baseline. 
The Euclidean distance improves $36.7\%$.
Introducing optimized weights improves the results of using the KL distance by $74.2\%$. 
Nevertheless, these classification procedures are outperformed by the maximum likelihood rule (ML), which is $87.3\%$ better than the baseline. 
Albeit remarkably good, this last result is overshadowed by our proposal that attains $100\%$ of accuracy, thus improving the baseline $100\%$. 

\begin{figure}[hbt]
\centering
\includegraphics[width=0.98\textwidth,height=0.93\textwidth]{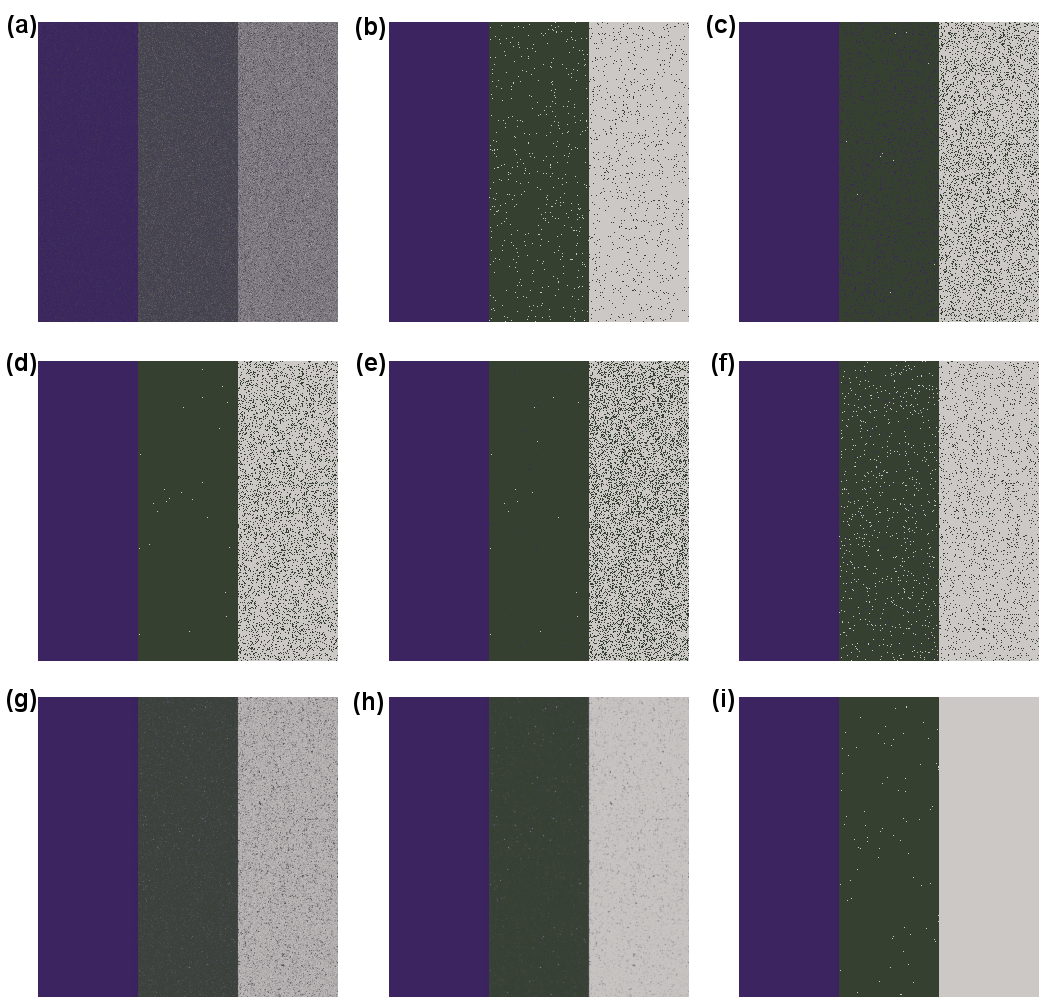}
\caption{Results of different classification techniques: (a)~original image, (b)~ML classification, (c)~ED classification, (d)~HD classification, (e)~KL classification, (f)~KL+OW classification, (g,h)~evolution of the solution of the diffusion-reaction system for $25$ and $50$ iterations respectively, and (i)~DR+KL+OW+$50$ classification. }
\label{fig:Phantom}
\end{figure}

Fig.~\ref{fig:Phantom} shows these results. 
We include two images that show the evolution of the solution of the diffusion-reaction equation for $n=25$ and $n=50$ iterations.  
Table~\ref{table:phantom} also shows the CPU time (in seconds) for each classification method. We use a simple \texttt{C++} implementation in an Intel Core~i5 (2.50~GHz) architecture without parallelization or other optimization techniques. 
We notice that, even with this basic implementation, the methods are quite fast and in particular the proposed method (DR+KL+OW+$50$) takes just $3.63$~seconds.

\subsection{Real Image}\label{Sec:ExperimentRealImage}

\begin{figure}[hbt]
\centering
\includegraphics[width=0.98\textwidth,height=0.72\textwidth]{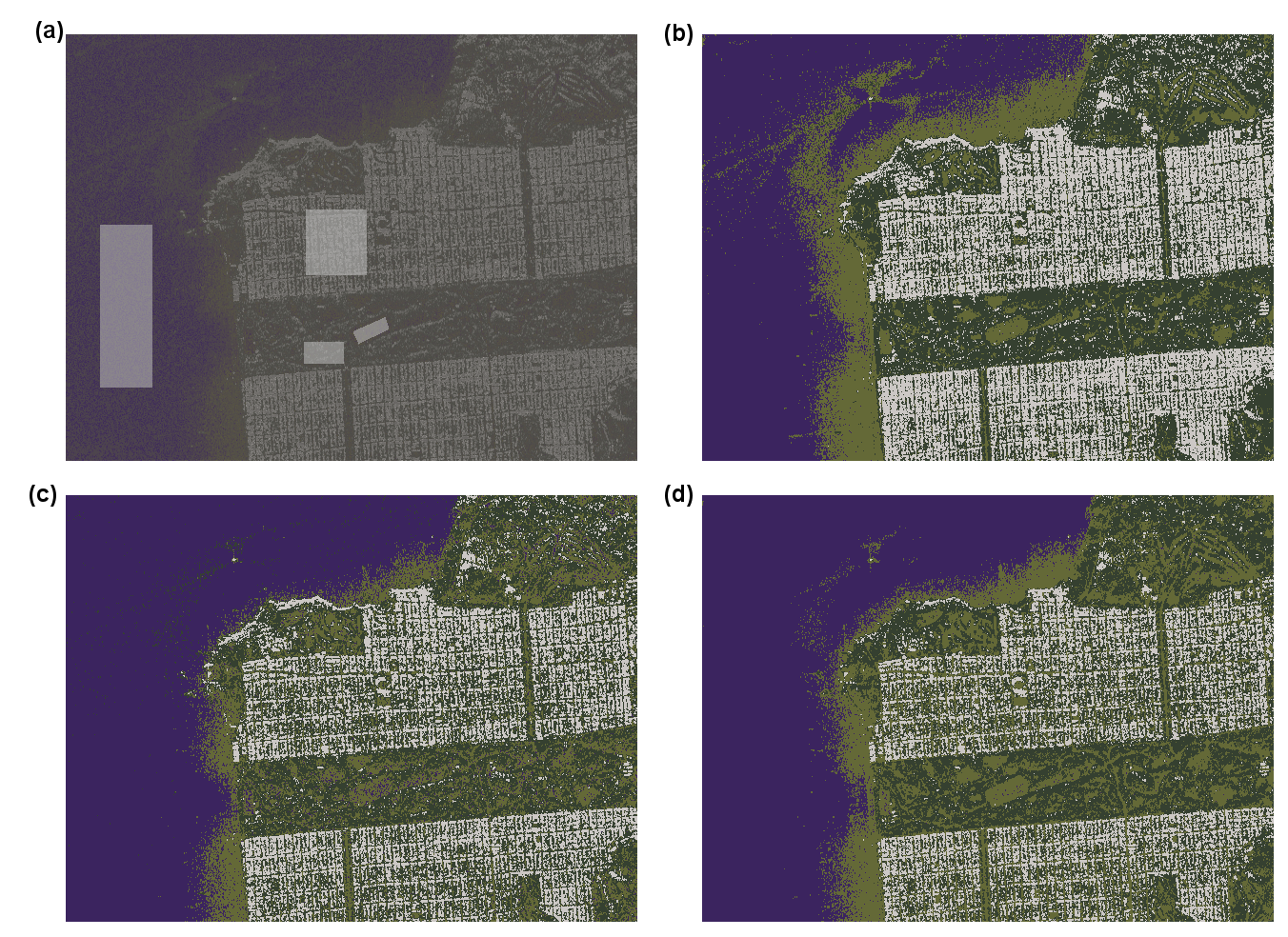}
\caption{(a)~San Francisco bay PolSAR image with ROIs, (b)~ML classification, (c)~ED classification, and (d)~HD classification.}
\label{fig:SF_01}
\end{figure}

In the second experiment we use a PolSAR image obtained by the AIRSAR sensor over the San Francisco bay.\footnote{Details about these freely available data can be accessed at ESA's web page \url{https://earth.esa.int/web/polsarpro/data-sources/sample-datasets}.} 
The image size is $600\times 448$ pixels. 
Four classes are readily identifiable: Ocean, Grass, Park and Urban areas. 
We selected a ROI for each class, whose observations were used to estimate the Wishart parameters according to the results presented in Section~\ref{correction:model}\ref{ref12}; these ROIs are shown in Fig.~\ref{fig:SF_01}. 
These data are also used as the reference to estimate the accuracy of the different classification strategies. 

\begin{figure}[hbt]
\centering
\includegraphics[width=0.98\textwidth,height=0.72\textwidth]{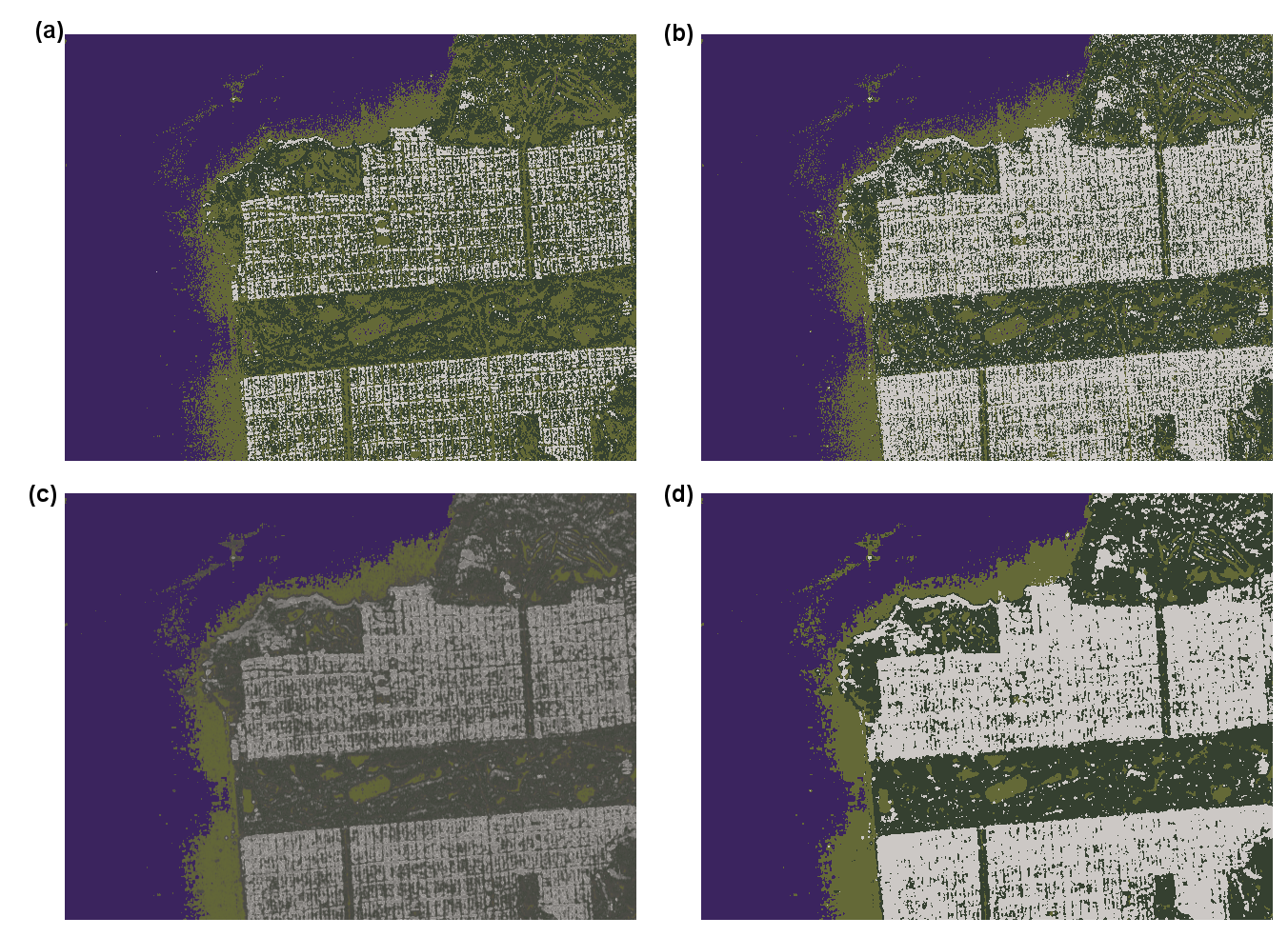}
\caption{(a)~KL classification, (b)~KL+OW classification , (c)~evolution of the solution of the diffusion-reaction system after $50$ iterations, and (d)~DR+KL+OW+$50$ classification. } 
\label{fig:SF_02}
\end{figure}

Table \ref{table:bay} presents the overall accuracy of classifying the four classes observed in the San Francisco image by the techniques here considered. 
With the exception of the Ocean class, that is almost perfectly classified by all methods, the worst result is used a
baseline.
Class-wide improvements with respect to the baseline are reported in parentheses. 

The results for the real image are consistent with the ones obtained for the simulated data: Euclidean and KL distances provide the worst results. 
Hellinger distance classification is the next best technique, except in the Urban area where the Euclidean distance is slighty  better. 
ML clearly outperforms Euclidean, KL and Hellinger distances, and even the weighted KL distance except in the case of Urban areas where the weighted KL distance is considerably better ($53.40\%$) than ML ($39.18\%$).
The optimized weights for this problem were $w=(0.44,0.38,0.12,0.06)$.
As expected, the Urban area presents a high variability which is consistent with the small weight value ($0.06$) obtained for this area in the optimization procedure. 

\begin{table}[hbt]
\centering
\begin{tabular}[c]{cr*4{l}}\toprule
\textbf{Method} & \multicolumn{4}{c}{\textbf{Accuracy (Improvement) by Class}} & 
\hspace{-3mm} \begin{tabular}{c} \textbf{CPU}\\ \textbf{time} \end{tabular}\\  
\cmidrule(lr{.75em}){2-5}
& Ocean & Grass & Park & Urban & \hspace{0mm} (s)\\ \midrule
ML & $99.8$ & $94.9$ ($82.5\%$) & $86.1$ ($60.1\%$) & $57.4$ ($39.3\%$) & 0.64 \\
ED & $99.8$ & $70.9$ (baseline) & $65.2$ (baseline) & $38.5$ ($12.4\%$) & 0.02\\
HD & $100.00$ & $93.7$ ($78.4\%$) & $68.7$ ($10.1\%$) & $34.8$ ($7.1\%$) & 0.81 \\
KL & $100.00$ & $89.4$ ($63.6\%$) & $65.4$ ($0.6\%$) & $29.8$ (baseline) & 0.09 \\
KL+OW & $100.00$ & $91.7$ ($71.5\%$) & $79.9$ ($42.2\%$) & $66.2$ ($51.9\%$) & 0.46\\
DR+KL+OW+$50$ & $100.00$ & $97.2$ ($90.4\%$) &  $93.8$ ($82.2\%$) & $84.4$ ($77.8\%$) & 7.67\\
\bottomrule
\end{tabular}
\caption{\label{table:bay} San Francisco bay classification scores using the test data.}
\end{table}

The proposed technique using the diffusion-reaction equation consistently improves in a significant way the classification accuracy ($90.4\%$, $82.2\%$ and $77.8\%$ for the Grass, Park and Urban areas, respectively), and when a perfect classification is obtained, as in Ocean, the iterative procedure does not make it worse. 
These improvements are not incremental with respect to both the baseline and to the use of optimized weights, and render classified images that can be considered excellent. 
In Fig.~\ref{fig:SF_01} and \ref{fig:SF_02} we illustrate the results of the different classification techniques for the San Francisco bay image. 
In table \ref{table:bay} we also show the CPU time for the different classification methods. The results are consistent with the ones obtained for the phantom image taking into account the size of the images. In particular the proposed method (DR+KL+OW+$50$) takes just $7.67$~seconds.  

\begin{figure}[hbt]
\centering
\includegraphics[angle=-90,width=.95\linewidth]{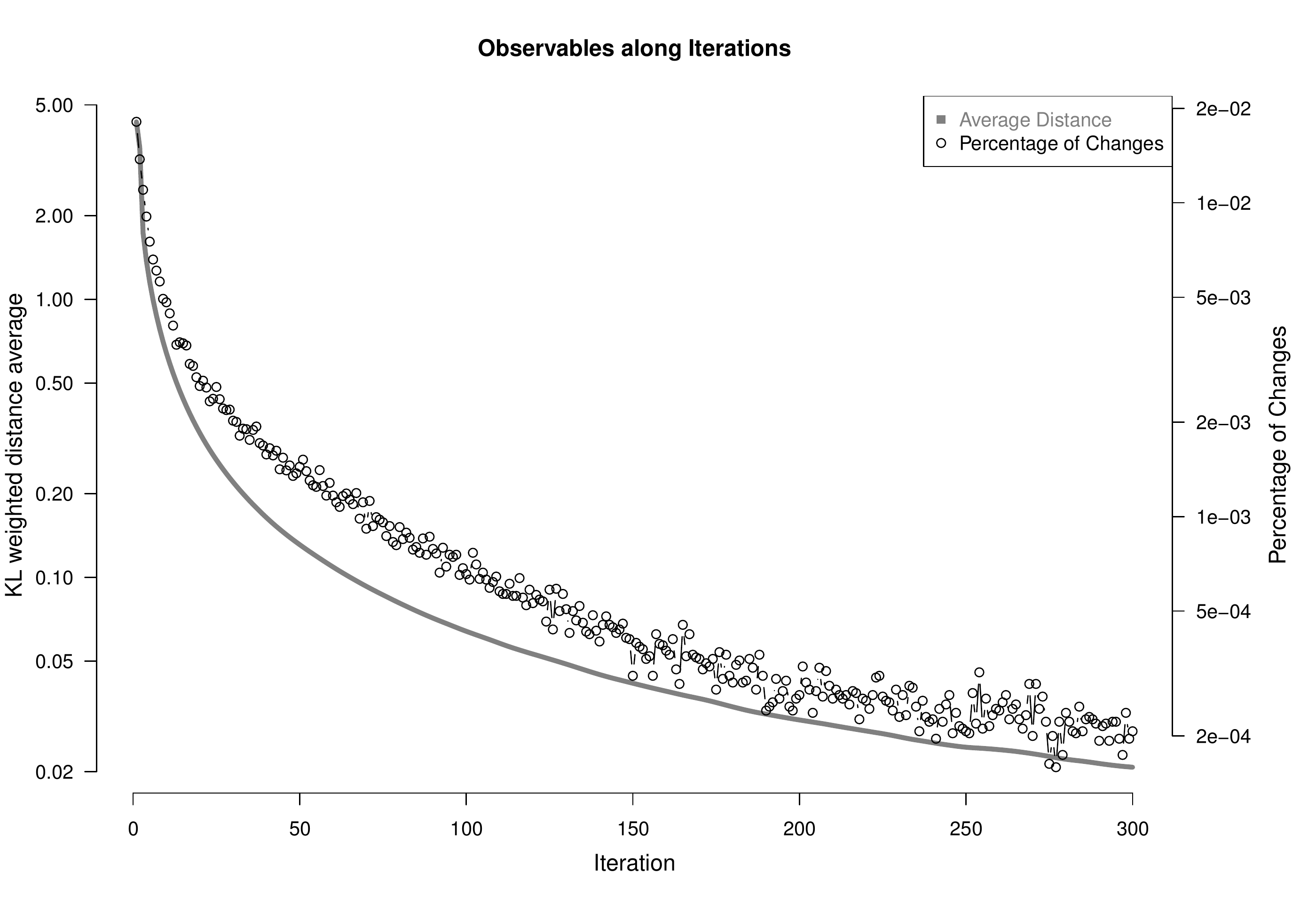}
\vspace{7mm}
\caption{Evolution of the diffusion-reaction system: KL weighted distance average of $\Sigma_{i,j}$ to its nearest class representative $\Sigma_{m_{\min}(\Sigma_{i,j})}$ (continuous gray curve), and percentage of pixels which change their class $m_{\min}(\Sigma_{i,j})$ (empty circles) in each iteration, semilogarithmic scale for the ordinates.} 
\label{fig:SF_RD_Evolution}
\end{figure}

Fig.~\ref{fig:SF_RD_Evolution} shows the evolution of the diffusion-reaction system described by Eq.~\eqref{PDE} observed in the classification of the San Francisco bay PolSAR image.
The ordinates of the two sequences are shown in logarithmic scale.
As explained above, the reaction term of the diffusion-reaction system tends to move each PolSAR image pixel value $\Sigma_{i,j}$ towards its nearest class representative  $\Sigma_{m_{\min}(\Sigma_{i,j})}$; in particular we expect that the weighted distance average of $\Sigma_{i,j}$ to $\Sigma_{m_{\min}(\Sigma_{i,j})}$ goes to zero when the data evolves according to the diffusion-reaction system. 
This behaviour is illustrated in the continuous gray curve.
Initially such distance average is equal to, approximately, $4.38$; 
we observe that it goes very quickly, faster than exponentially, to zero. 
In a complementary fashion, the diffusion term  tends to smooth the solution and, as a consequence, it makes some pixels change their classification.
The circles show the evolution of the percentage of pixels which change their associated class in each iteration (that is,  the value  $m_{\min}(\Sigma_{i,j})$ is altered). 
We observe that $1.9\%$ of pixels change their associated class in the first iteration, and that this rate decreases across the iterations also faster than exponentially.

\section{Conclusions}\label{Sec:Conclusion}

In this paper we address the problem of PolSAR image classification. 
The main contributions of the paper are, on the one hand to introduce a stochastic weighted distance strategy to increase the class discrimination power and on the other hand to embed the classification in a diffusion-reaction differential system which aims to include smoothing constraints in the classification procedure. 
The weights used to improve the class discrimination power of the distance are estimated by minimizing a new energy functional. 
These weights take into account the PolSAR matrix variability in the class regions (the lower the weight, the larger the expected variability inside the class). 

We show that introducing these weights is equivalent to modifying the number of looks of the Wishart model for each class and, as discussed by Torres et al.~\cite{art:Torres_2014}, this takes account of the variability due to texture that is not embedded in the Wishart model.
We present experiments with simulated and real PolSAR images.
These experiments confirm that the introduction of the weighted distances improves considerably the classification results. 

By embedding the classification problem in a diffusion-reaction differential system, and using the weighted distances, the classification obtained with simulated and real data outperforms, in a significant way, the classification score obtained with the usual classification strategies including maximum likelihood, and Euclidean, Hellinger and Kullback-Leibler distances.

Further comparisons, including computational costs, will be made with respect to other contextual techniques as, for instance, the simple mode and the ICM algorithm~\cite{FreryCorreiaFreitas:ClassifMultifrequency:IEEE:2007}.
Also, techniques for estimating optimal values of the parameters that govern the process ($\alpha$ and $\delta t$) will be sought.

\bibliographystyle{vancouver}
\bibliography{Speckle_bibliography}

\end{document}